\UseRawInputEncoding
\pdfoutput=1 
\documentclass[sigconf]{acmart}

\AtBeginDocument{%
  \providecommand\BibTeX{{%
    \normalfont B\kern-0.5em{\scshape i\kern-0.25em b}\kern-0.8em\TeX}}}

\setcopyright{acmcopyright}
\copyrightyear{2018}
\acmYear{2018}
\acmDOI{XXXXXXX.XXXXXXX}

\acmConference[Conference acronym 'XX]{Make sure to enter the correct
  conference title from your rights confirmation emai}{June 03--05,
  2018}{Woodstock, NY}
\acmPrice{15.00}
\acmISBN{978-1-4503-XXXX-X/18/06}

\usepackage{amssymb}
\usepackage{tikz}
\usepackage{multirow}
\acmSubmissionID{xxxx}

\begin{document}

\title{TextCLIP: Text-Guided Face Image Generation And Manipulation Without Adversarial Training}


\author{Xiaozhou You}
\acmSubmissionID{xxxx}
\affiliation{
  \institution{School of Electronic and Computer Engineering, Peking University}
  \country{China}
  }

\author{Jian Zhang}
\acmSubmissionID{xxxx}
\affiliation{
	\institution{School of Electronic and Computer Engineering, Peking University}
	\country{China}}

\settopmatter{printacmref=false}
\renewcommand\footnotetextcopyrightpermission[1]{}

\pagestyle{empty}
\begin{abstract}
  Text-guided image generation aimed to generate desired images conditioned on given texts, while text-guided image manipulation refers to semantically edit parts of a given image based on specified texts. For these two similar tasks, the key point is to ensure image fidelity as well as semantic consistency. Many previous approaches require complex multi-stage generation and adversarial training, while struggling to provide a unified framework for both tasks. In this work, we propose TextCLIP, a unified framework for text-guided image generation and manipulation without adversarial training. The proposed method accepts input from images or random noise corresponding to these two different tasks, and under the condition of the specific texts, a carefully designed mapping network that exploits the powerful generative capabilities of StyleGAN and the text image representation capabilities of Contrastive Language-Image Pre-training (CLIP) generates images of up to $1024\times1024$ resolution that can currently be generated. Extensive experiments on the Multi-modal CelebA-HQ dataset have demonstrated that our proposed method outperforms existing state-of-the-art methods, both on text-guided generation tasks and manipulation tasks.
\end{abstract}





\begin{teaserfigure}
\setlength{\abovecaptionskip}{0.1cm}
  \includegraphics[width=\linewidth]{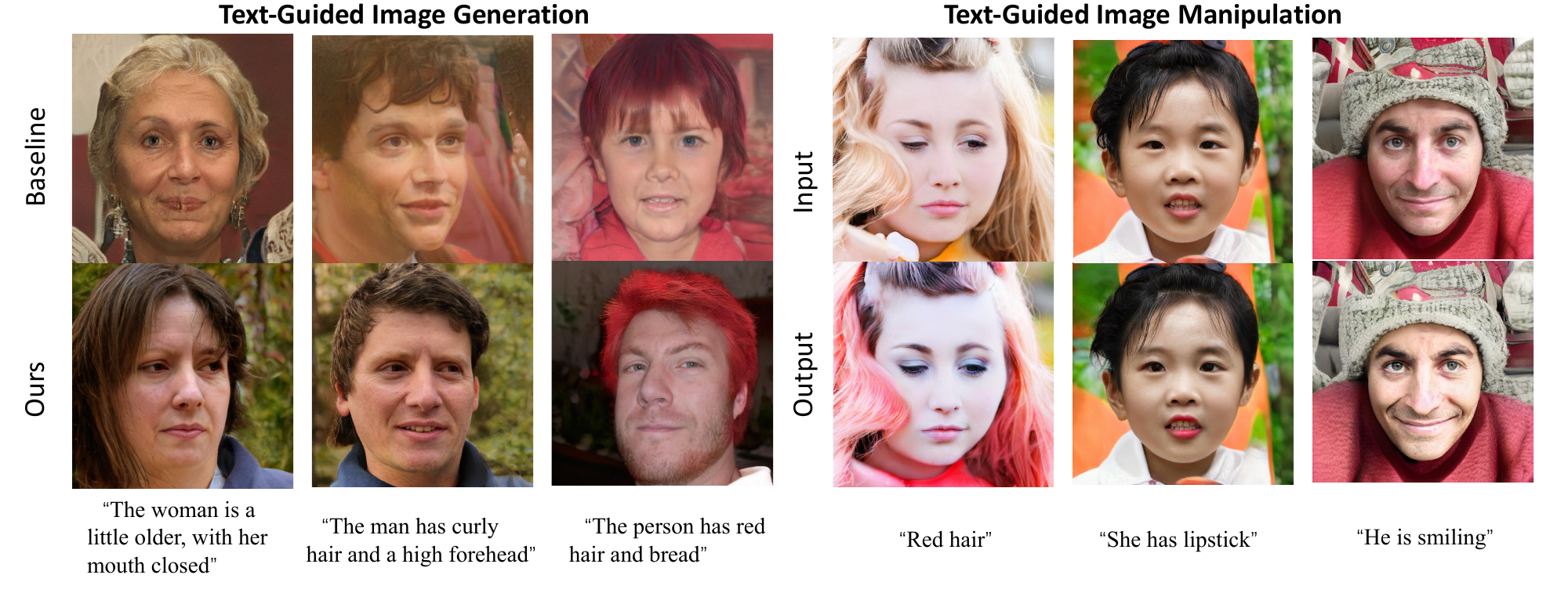}
  \caption{In this work, we propose TextCLIP, a unified framework for text-guided image generation and manipulation without adversarial training. On the left is the comparison between TextCLIP and baseline \cite{xia2021tedigan} on image generation task, on the right are the results of TextCLIP on image manipulation task.}
  \label{f1}
  \vspace{0.2cm}
\end{teaserfigure}
\maketitle

\section{Introduction}
Text-guided image generation and manipulation has recently gained significant attention and made some progress in the field of computer vision \cite{zhang_cross-modal_2021,xia2021tedigan,patashnik2021styleclip,ding2021cogview,ramesh_zero-shot_2021}. Text-guided image generation and manipulation require the generation or modification of images based on specified text, which is two complex cross-modal tasks. Text and images belong to two different modalities, and cross-modal data operation is difficult. 
\begin{table*}[t]
	\caption{Comparison of Different Text-Guided Image Generation Models.}
	\label{t1}
	\begin{tabular}{ c | cccccc }
		\toprule
		Method  & AttnGAN \cite{xu_attngan_2018}  &  ControlGAN \cite{li2019controllable} &  DAE-GAN \cite{ruan_dae-gan_202}&XMC-GAN \cite{zhang_cross-modal_2021} & TediGAN \cite{xia2021tedigan} & \textbf{TextCLIP}  \\ 
		\midrule
		One Generator&-&- &-&\checkmark&\checkmark&\checkmark \\ 
		Single Model&\checkmark&\checkmark&\checkmark&\checkmark&-&\checkmark\\ 
		High Resolution&-&-&-&-&\checkmark&\checkmark\\
		Manipulation&-&\checkmark&-&-&\checkmark&\checkmark\\
		Open World &-&-&-&-&\checkmark&\checkmark\\
		w/o Adversarial Training&-&-&-&-&\checkmark&\checkmark\\ 
		\bottomrule 
	\end{tabular}
\end{table*}
For the task of text-guided image generation, Reed \emph{et al.} \cite{reed_generative_2016} first proposed text-guided image generation using adversarial generative networks \cite{goodfellow_generative_2014} and generated more research on text-guided image generation \cite{zhang_stackgan_2017,zhang_stackgan_2018,xu_attngan_2018,zhu_dm-gan_2019,li2019controllable,zhang_cross-modal_2021,wang2021cycle,zhu2020cookgan,gou2020segattngan}. The generated image needs to not only produce a sufficiently realistic image, but also be semantically consistent with the corresponding text. Some previous research has focused on multi-stage generation, where multiple low-quality images are first generated to produce high-quality images, which means that multiple generators and discriminators need to work together. These efforts require a tedious multi-stage generation process and complex adversarial training, which is very time-consuming and difficult to train. For other recent works \cite{xia2021tedigan,zhang_cross-modal_2021,ramesh_zero-shot_2021,ding2021cogview}, some of them have much room for improvement in the quality of the generated images, while the others require a large number of training parameters or training data, making training too expensive. 

For text-guided image manipulation task \cite{jiang2021talk,zhou2021generative,sun2021multi,hou2022textface,cherepkov2021navigating}, the corresponding image needs to be modified according to the specified text. It is important to note that the areas of the image that are semantically irrelevant to the specified text should be kept as close as possible to the original image, and only those areas of the image that are semantically relevant should be modified.TediGAN \cite{xia2021tedigan} is the first work to provide text-guided image generation and manipulation by exploiting the semantic properties of the latent space of GAN. However, the performance of TediGAN has much room for improvement. 

StyleGAN \cite{karras2019style,karras_analyzing_2020,karras2020training,karras2021alias} is now state-of-the-art generative adversarial networks with powerful image generation capabilities, providing realistic images at resolutions up to $1024\times1024$, and more importantly, StyleGAN's latent space with good semantic performance and unraveling capabilities. The latent space of StyleGAN has been the subject of much recent research progress \cite{wu_stylespace_2021,shukor2021semantic,abdal_image2stylegan_2019}, which has significantly advanced several fields. Contrastive Language-Image Pre-training (CLIP) \cite{radford2021learning} is a powerful multimodal pretrained model that provides powerful text image representation capabilities and can be used as a supervisor for cross-modal tasks to achieve semantic-visual alignment. Some meaningful works based on pretrained StyleGAN and CLIP have been born recently \cite{patashnik2021styleclip,alaluf_restyle_2021,xia_towards_2021,pakhomov2021segmentation,chefer2021image,gao2021clip,schaldenbrand2021styleclipdraw,zhang2021dance}.

In this work, we propose TexCLIP, a unified framework for text-guided image generation and manipulation without adversarial training, which doesn't require the complex multi-stage generation and tedious adversarial training and outperforms extant state-of-the-art methods in two tasks. First, either random noise or images are used as input, with the random noise corresponding to the text-guided image generation task and the images corresponding to the text-guided image manipulation. Using a pre-trained encoder, the input is transformed into $w_{0}$, which is used as the initial latent code. $w_{0}$ is then subjected to a level-channel mapper with two parts: (a) level mapper: from coarse to fine, divided into three separate networks (coarse, medium, fine), each mapping a part of the initial latent code $w_{0}$. (b) channel mapper: consists of 18 style modulation networks. The final mapping latent code $w_{t}$ is obtained by level-channel mapper, which is then processed differently with the initial latent code $w_{0}$  for different tasks to obtain the final latent code $w_{s}$. $w_{s}$ is used as input to the generator of StyleGAN to obtain the final image. Table \ref{t1} shows how our method compares with other methods. Compared with other text-guided image generation methods, our proposed method is able to produce high-resolution images, support manipulation of images and accept open-world text as input without the need for adversarial training and multi-stage generation. In contrast to TediGAN \cite{xia2021tedigan}, we do not need to train different models for different texts.

In summary, this work consists of the following main contributions:
\begin{itemize}
    \item For the two distinct tasks of text-guided image generation and text-guided image manipulation, we propose TextCLIP, a unified framework that enables text-guided image generation and manipulation without the need for complex adversarial training.
    \item We propose level-channel mapper that uses text as a condition to semantically map the initial latent code to the latent space $\mathcal{W}+$ \cite{abdal_image2stylegan_2019} of StyleGAN. Compared to previous work TediGAN \cite{xia2021tedigan}, level-channel mapper does not require training different networks for different text conditions.
    \item Extensive qualitative and quantitative studies have shown that our proposed TextCLIP outperforms existing state-of-the-art methods on these two different tasks.
\end{itemize}

\section{Related Work}
\subsection{Text-Guided Image Generation}
We have divided previous work on text-guided image generation into two categories. The first category is multi-stage generative models, where multiple generators and discriminators need to be used to complete the text-guided image generation work. StackGAN \cite{zhang_stackgan_2017} was the first multi-stage generative model that used multiple generators and discriminators to first generate low-quality images and then generate high-quality images. Later StackGAN++ \cite{zhang_stackgan_2018} implemented end-to-end training based on StackGAN to generate higher quality images. AttnGAN \cite{xu_attngan_2018} introduced an attention mechanism to achieve word-level image generation, generating more realistic and realistic high-quality images; in addition, the proposed Deep Attention Multimodal Similarity Model (DAMSM) to compute the similarity of image-text pairs. DM-GAN \cite{zhu_dm-gan_2019} ugenerates a low-resolution initial image with a smaller model size, and then uses dynamic memory networks to purify the initial image to produce a more realistic image. Much subsequent work, optimised on the basis of AttnGAN, has achieved higher quality image generation with more accurate semantic alignment \cite{li2019controllable,qiao_mirrorgan_2019,cheng_rifegan_2020,ruan_dae-gan_202}.ControlGAN \cite{li2019controllable} proposes an innovative multi-stage generation architecture and introduces perceptual loss to solve the problem that if some words in a sentence are changed during text-guided image generation, the composite image will be very different from the original image.MirrorGAN \cite{qiao_mirrorgan_2019} is inspired by CycleGAN \cite{zhu2017unpaired} and reduces the generated images to text, further improving the quality of the generated images. DAE- GAN \cite{ruan_dae-gan_202} takes into account the 'aspect' information of the input text and incorporates it into the multi-stage generation process.

The second category is represented by XMC-GAN \cite{zhang_cross-modal_2021} and DALL-E \cite{ramesh_zero-shot_2021}. XMC-GAN \cite{zhang_cross-modal_2021} uses contrast learning as supervision, takes into account image text contrast loss, true-false image contrast loss, and image region word contrast loss, and uses modulation layers to build a single-stage generative network that achieves state-of-the-art performance on several public dataset. DALL-E \cite{ramesh_zero-shot_2021} trains a large number of text-image pairs on a Transformer with 12 billion network parameters, achieving zero-s
hot generation. CogView \cite{ding2021cogview} is similar to DALL-E in that it trains a Transformer \cite{vaswani2017attention} with 4 billion network parameters to autoregressively model images and text, achieving stronger zero-sample generation. TediGAN-A \cite{xia_gan_2021} uses a pretrained StyleGAN with a GAN inverse module, a visual semantic similarity module, and an instance-level optimization module to perform an optimized search in the latent space, resulting in text-guided image generation. TediGAN-B \cite{xia_towards_2021} improves the performance of TediGAN-A by using a pretrained image inversion model and CLIP \cite{radford2021learning}.
\begin{figure}[t]
\setlength{\abovecaptionskip}{0cm}
  \includegraphics[width=\linewidth]{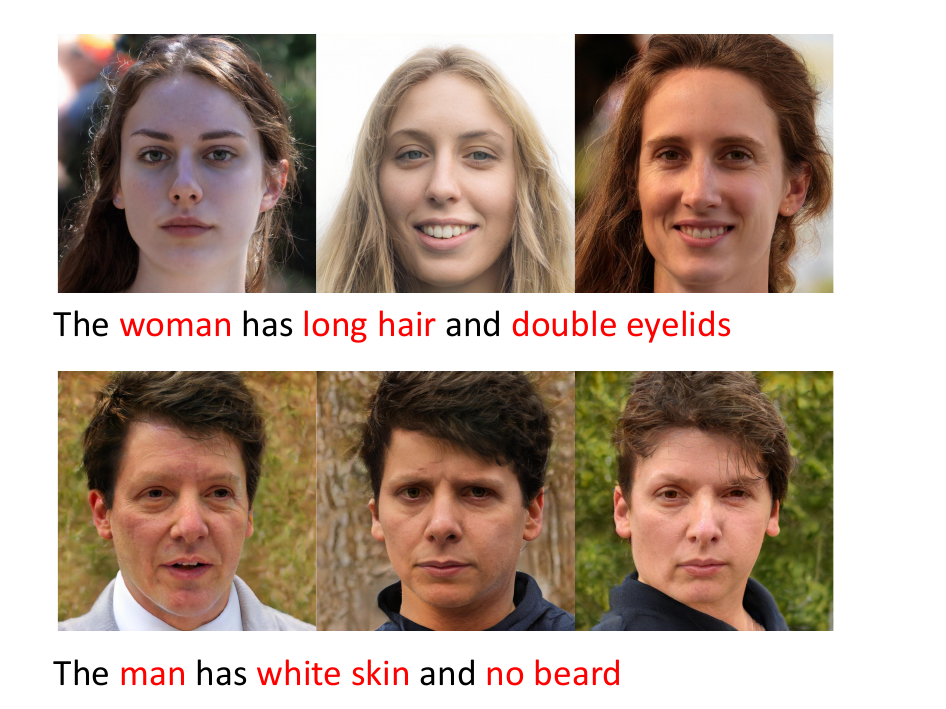}
  \caption{Diverse text-guided image generation results. On the same text conditions, TextCLIP can generate multiple images at $1024\times1024$ resolution.}
  \label{ff2}
\end{figure}
\subsection{Text-Guided Image Manipulation}
ManiGAN \cite{li2020manigan} is a multi-stage text-guided image manipulation work using multiple generators and discriminators and has demonstrated good performance on the CUB and COCO datasets. StyleCLIP \cite{patashnik2021styleclip} provides three different methods for text-guided image manipulation, including optimizers, mappers and global direction. The mapper requires training a model with different parameters for different text conditions and is an inflexible approach for practical applications. The optimizer and global direction approaches require inference on different instances each time and take longer to infer. Our proposed TextCLIP differs from previous work in that we propose a more flexible way to perform text-guided image manipulation and achieve higher quality image manipulation. Instead of training a different model for each text, TextCLIP can use the trained model to generate results directly based on the image and text conditions, without excessive inference time.For example, we can train a model for the same class of text conditions, e.g. a model trained about skin color can perform inference on dark skin, white skin, red skin, etc.
\begin{figure*}[!t]
\setlength{\abovecaptionskip}{0.15cm}
  \includegraphics[width=\textwidth]{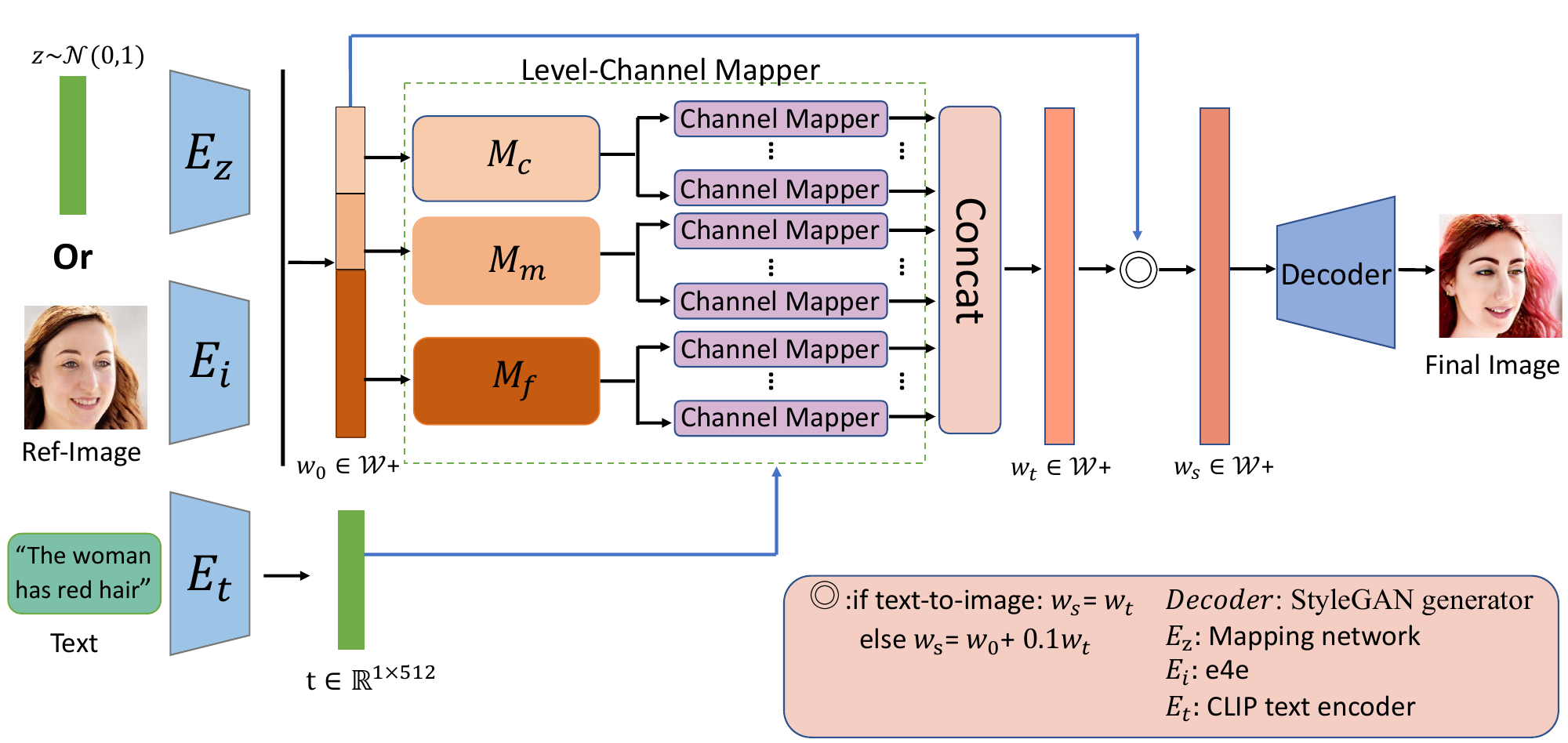}
  \caption{The framework of TextCLIP. The level mappers $M_{c},M_{m},M_{f}$ consist of several fully connected layers that take a part of $w_{0}$ as input. There are a total of 18 channel mappers, each taking $t$ encoded by the CLIP text encoder and the output of the corresponding level mapper as input. The outputs of the 18 channel mappers are concatenated to form $w_{t}$.$w_{t}$ is then processed differently for different tasks to obtain $w_{s}$. $w_{s}$ is used as input of the pretrained StyleGAN generator to obtain the final image.}
  \label{f2}
\end{figure*}
\subsection{StyleGAN And CLIP}
StyleGAN \cite{karras2019style,karras_analyzing_2020,karras2020training,karras2021alias} is an excellent tool for image generation and is state-of-the-art work in the field of adversarial generative networks. StyleGAN's input is mapped to the latent space by processing eight fully connected layers, which are then fed into the StyleGAN generator. The StyleGAN generator has 18 layers, with every two layers corresponding to a resolution from 2 to 1024. Each layer of the generator of StyleGAN accepts a 512-dimensional latent code as input. Due to the good semantic properties of the latent space of StyleGAN, many extensions on the latent space of StyleGAN have been born recently, such as $\mathcal{W}+$ and $\mathcal{S}$ space, and these researches are good to enhance the applications of StyleGAN. The $\mathcal{W}+$ space of StyleGAN consists of 18 512-dimensional latent codes, each corresponding to one of the layers of StyleGAN generator and serving as its input. The excellent performance of StyleGAN's $\mathcal{W}+$ space has driven advances in the field of GAN inversion. GAN inversion work \cite{tov_designing_2021,richardson2021encoding,huh2020transforming,viazovetskyi2020stylegan2,zhu2020cookgan,esser2020disentangling,collins2020editing} can well invert images into the $\mathcal{W}$+ space of StyleGAN, thus facilitating semantic editing of images. Contrastive Language-Image Pre-training (CLIP) \cite{radford2021learning} trains a large number of image-text pairs, providing a powerful image-text representation. By encoding the image and text into the space of CLIP, the similarity of the image text can be quantified.

\section{The TextCLIP Framework}
Based on the powerful image generation capability of StyleGAN \cite{karras_analyzing_2020} and the cross-modal text-image representation capability of CLIP \cite{radford2021learning}, we propose TextCLIP, a unified approach for text-guided image generation and manipulation. We divide TextCLIP into three stages:
\begin{itemize}
    \item \textbf{Stage 1.} Using a pretrained encoder, the image or random noise is mapped to the $\mathcal{W}+$ \cite{abdal_image2stylegan_2019} space of StyleGAN model pretrained on the FFHQ dataset \cite{karras2019style} to obtain an initial latent code $w_{0}$.
    \item \textbf{Stage 2.} The initial latent code $w_{0}$ is passed through the level-channel mapper to obtain the mapping latent code $w_{t}$. 
    \item \textbf{Stage 3.} The mapping latent code $w_{t}$ is then processed differently with the initial latent code $w_{0}$ depending on the task to obtain the style latent code $w_{s}$, which is the input of the generator of a pretrained StyleGAN to obtain the final image.
\end{itemize}

\subsection{Overview}
The global framework is shown in Figure \ref{f2}. TextCLIP supports either random noise or image as input (random noise for text-guided image generation task and image for text-guided image manipulation task), and we use a pretrained encoder to map the input to the latent space $\mathcal{W}+$ of StyleGAN\footnote{In our experiments, we actually use StyleGAN2 \cite{karras_analyzing_2020}.}. For image, we use e4e \cite{tov_designing_2021} as the pretrained encoder; for random noise, we use a pretrained mapping network in StyleGAN \cite{karras_analyzing_2020} as encoder. the process can be formulated as:
\begin{equation}
  w_{0}=E(g_{0}),
\end{equation}
where $g_{0}$ represents the initial input,$w_{0}$ represents the initial latent code mapped to the $ \mathcal{W}+ $ space of StyleGAN and $E$ represents the pretrained encoder. The obtained initial latent code is then passed through the level-channel mapper to obtain the mapping latent code $w_{t}$, the mathematical equation of which is shown below:
\begin{equation}
  w_{t}=F_{LCM}(w_{0}),
\end{equation}
where $F_{LCM}$ denotes the level-channel mapper. Next, we do different things depending on the task. For text-guided image generation task:
\begin{equation}
  w_{s}=w_{t},
\end{equation}

For text-guided image manipulation task:
\begin{equation}
  w_{s}=0.1w_{t} +w_{0},
\end{equation}
where the style latent code $w_{s}$ is used as the input of StyleGAN generator to obtain the final image $g_{s}$. The mathematical equation is shown below:
\begin{equation}
  g_{s}=G(w_{s}),
\end{equation}
where $G$ denotes the generator of a pretrained StyleGAN, $w_{0}$, $w_{t}$ and $w_{s} \in \mathcal{W}+$. 

\subsection{Level-Channel Mapper}
The level-channel mapper consists of two parts: the level mapper and the channel mapper.

\subsubsection{\textbf{Level Mapper}}

Many previous studies have shown that different layers of StyleGAN generator control different attributes, so from coarse to fine we divided the layers of StyleGAN generator into three parts (coarse, medium, fine). In the same way we divided the input latent code $w_{0}$ into three parts, as follows: 
\begin{equation}
  w_{0}=(w_{0}^{c},w_{0}^{m},w_{0}^{f}),
\end{equation}

For each part, we design a network consisting of several fully connected layers, each of which is followed by operations such as layernorm and leaklyrelu. This is shown below:
\begin{equation}
  M(w_{0})=(M^{c}(w_{0}^{c}),M^{m}(w_{0}^{m}),M^{f}(w_{0}^{f})).
\end{equation}

In practice, we can train only one sub-network of $M$. Doing so allows us to change only the relevant image attributes and not some irrelevant ones.

As shown in Table \ref{tt}, experimental results show that each layer of StyleGAN \cite{karras_analyzing_2020} controls different attributes, such as eye, hair color, age, face color, and other attributes. After our division, the coarse level controls attributes such as nose, head shape, lips, and hair length; the middle level controls attributes such as hair and face color; and the fine level controls age, gender and some micro attributes.
\subsubsection{\textbf{Channel Mapper}}
\begin{figure}[t]
  \includegraphics[width=\linewidth]{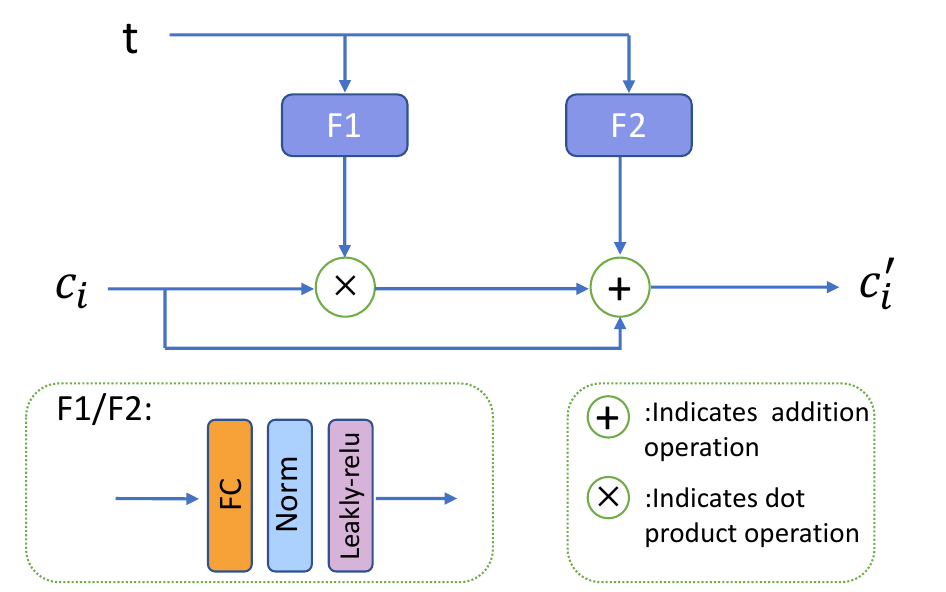}
  \caption{The structure of channel mapper.$t$ is the text vector encoded by the CLIP text encoder. $c_{i}$ is the input of the ith channel mapper and comes from the corresponding level mapper.}
  \Description{Enjoying the baseball game from the third-base
  seats. Ichiro Suzuki preparing to bat.}
  \label{f4}
\end{figure}

We design a channel mapper for each layer of a StyleGAN generator. There are 18 channel mappers in total. $M^{c}$ corresponds to 4 channel mappers, $M^{m}$to 4 channel mappers and $M^{f}$ to 10 channel mappers. For each channel mapper, it takes the output from the corresponding level mapper and the text code $t$ encoded by the CLIP \cite{radford2021learning} text encoder as input. As shown in Figure \ref{f4}, the text is first encoded by CLIP text encoder to obtain text conditional code $t$. $t$ modulates the input that comes from the corresponding level mapper after processing in two fully connected layers. The mathematical form is shown below:
\begin{equation}
  c_{i}'=c_{i}+F_{1}(t) c_{i} + F_{2}(t),i=0,1,...,17 ,
\end{equation}
where $F_{1}$ and $F_{2}$ are two networks designed by fully connected layers,$c_{i}$ is the input of layer $i$. Finally, the resulting 18 channel styles are concatenated to obtain the final style latent code $w_{s}$. The mathematical form shown below:
\begin{equation}
  w_{s}=Concat(c_{0}' ,c_{1}' ,c_{2}',...,c_{17}'),
\end{equation}
where $Concat$ means that the outputs of the 18 channel mappers are sequentially concatenated together.
\begin{figure*}[!t]
\setlength{\abovecaptionskip}{0.15cm}
  \includegraphics[width=\textwidth]{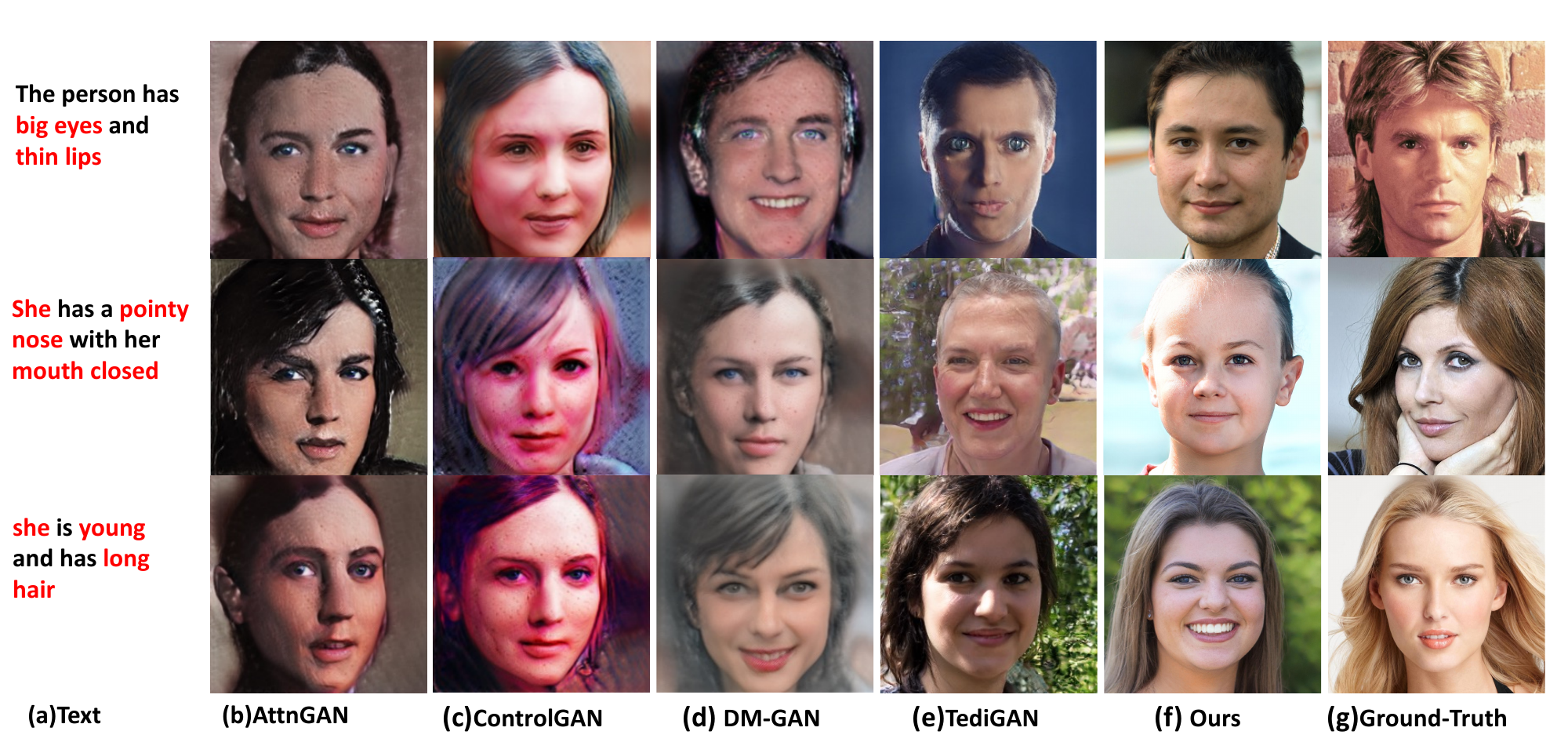}
  \caption{Qualitative comparison of text-gudied image generation compared with the state-of-the-art methods. TextCLIP generates more realistic and semantically similar images than previous methods.}
  \label{f5}
\end{figure*}

\subsection{Loss Function}
\subsubsection{\textbf{Semantic Loss}}
An important aspect of text-guided image generation and manipulation is the need to ensure that the generated images are semantically consistent with the corresponding text. For this consideration, we propose semantic loss. The text and image are first encoded separately using CLIP \cite{radford2021learning} pretrained encoder, and then the result is computed as the cosine similarity to obtain the semantic loss. 
\begin{equation}
  \mathcal{L}_{semantic}=1-cos(t,F_{img}(G(w_{s}))),
\end{equation}
where $F_{img}$ represents the pretrained image encoder of CLIP, $t$ is the text vector obtained by processing the CLIP text encoder, and $cos$ represents the cosine similarity calculation, $\mathcal{L}_{semantic}$ represents semantic loss.
\subsubsection{\textbf{Identity Loss}}
We need to ensure that the generated image is identical to the original facial identity, so we introduce identity loss as follows:
\begin{equation}
  \mathcal{L}_{ID}=1-cos(R(g_{0}),(R(G(w_{s})))),
\end{equation}
where $g_{0}$ represents the original image and $R$ represents a pretrained Arcface \cite{deng2019arcface} network for extracting the identity features of the image. The identity loss $ \mathcal{L}_{ID} $ is obtained by calculating the cosine similarity of the face identity features of the two images.
\subsubsection{\textbf{Image Loss}}
The image loss consists of pixel loss $ \mathcal{L}_{pixel} $ and image feature loss $ \mathcal{L}_{lpips}$. Pixel loss refers to the fine-grained supervision of the generated image by comparing each pixel of the generated image with the original image $g_{0}$. Feature loss refers to the comparison of the images at the feature level, typically using a pretrained network for feature extraction \cite{johnson2016perceptual}. Image loss is defined as follows:
\begin{equation}
  \mathcal{L}_{pixel}=\Vert g_{0}-G(w_{s}) \Vert_{2}^{2},
\end{equation}
\begin{equation}
  \mathcal{L}_{lpips}=\Vert F_{VGG}(g_{0})-F_{VGG}(G(w_{s}))  \Vert_{2}^{2},
\end{equation}
where $F_{VGG}$ represents a pretrained VGG network for extracting image features \cite{johnson2016perceptual}. The total image loss is shown below:
\begin{equation}
  \mathcal{L}_{img}=\lambda_{pixel}\mathcal{L}_{pixel} +\lambda_{lpips}\mathcal{L}_{lpips} ,
\end{equation}
where $\lambda_{pixel},\lambda_{lpips}$ are the corresponding hyperparameters.

\subsubsection{\textbf{Fidelity Loss}}
After experimentation, it was found that previous research in text-guided image generation and manipulation tended to produce some low-quality and blurred images. To address this issue, we introduce the fidelity loss to prevent the generation of some low-quality and blurred images. It is shown as follows:
\begin{equation}
  \mathcal{L}_{d}=\sigma(D(g_{s}))),
\end{equation}
where $\sigma$ represents sigmoid function, $g_{s}$ represents generated image, $D$ represents StyleGAN discriminator. We use a pretrained discriminator $D$ of StyleGAN \cite{karras_analyzing_2020}, which performs image fidelity determination to prevent the model from generating blurred photos.
\subsubsection{\textbf{Overall Loss}}
In summary, in order to make the images generated by the model realistic and semantically similar to the corresponding text, we define the following loss function:
\begin{equation}
  \mathcal{L}=\lambda_{semantic}\mathcal{L}_{semantic} +\lambda_{ID}\mathcal{L}_{ID} +\lambda_{img}\mathcal{L}_{img} + \lambda_{d}\mathcal{L}_{d} ,
\end{equation}
where $ \lambda_{semantic},\lambda_{ID},\lambda_{img},\lambda_{d}$ are the corresponding hyperparameters.

\begin{table}[t]
  \caption{Layer-wise Analysis of a 18-layer StyleGAN Generator.}
  \label{tt}
  \begin{tabular}{c|c|l}
    \toprule
    Level&Layers&Attributes\\
    \midrule
    coarse & 0-3& face shape,hair length,nose,lip,\emph{et,al.}\\
    medium & 4-7& hair color,face color,\emph{et,al.}\\
    fine & 7-17& age,gender,micro features,\emph{et,al.}\\
  \bottomrule
\end{tabular}
\end{table}
\section{Experiments}
\subsection{Experiments Setup}

\subsubsection{\textbf{Datasets}}
In order to carry out the performance of text-guided face image generation and manipulation, we conducted our experiments to verify the soundness and efficiency of the TextCLIP method. We have selected the following face dataset to carry out our experiments.
\begin{itemize}
    \item \textbf{Multi-modal CelebA-HQ Dataset} \cite{xia2021tedigan}: a multimodal dataset consists of images, descriptive text, semantic masks and sketch,and contains 30,000 images, 24,000 images in the training set and 6,000 images in the test set. Each image of Multi-modal CelebA-HQ Dataset corresponds to 10 text descriptions.
\end{itemize}

\subsubsection{\textbf{Evaluation Metric}}
Text-guided image generation and manipulation require that the generated images are not only really enough to be realistic but also maintain a semantic similarity to the corresponding text. For this purpose, we have chosen the following evaluation metric.
\begin{itemize}
    \item \textbf{Frechet Inception Distance (FID)} \cite{heusel2017gans}: FID represents the distance between the feature vectors of the generated image and the feature vectors of the real image. The closer the distance is, the better the result of the model.FID gives us a good indication of whether the model is generating the exact data we desired.
    \item \textbf{R-Precision} \cite{xu_attngan_2018}: another important property of text-guided image generation and manipulation is semantic similarity.we use R-Precision which evaluates the top-1 retrieval accuracy as the major evaluation metric in an image The higher the value of R-Precision, the higher the semantic similarity.
    \item L\textbf{earned Perceptual Image Patch Similarity(LPIPS)} \cite{zhang2018unreasonable}: to further evaluate the similarity of the generated image and the original image, we use LPIPS, which is a metric that learns the inverse of the generated image and the real image. A lower value of LPIPS indicates that the two images are more similar.
    \item \textbf{Identity similarity(IDS)} \cite{huang2020curricularface}: for text-guided image manipulation, we want the modified face image to be identity consistent with the original image, so we use IDS to evaluate this performance. IDS denotes identity similarity before and after editing calculated by Curricularface. The higher the IDS, the better the identity similarity.
\end{itemize}
\begin{figure*}[!t]
\setlength{\abovecaptionskip}{0.15cm}
  \includegraphics[width=\textwidth]{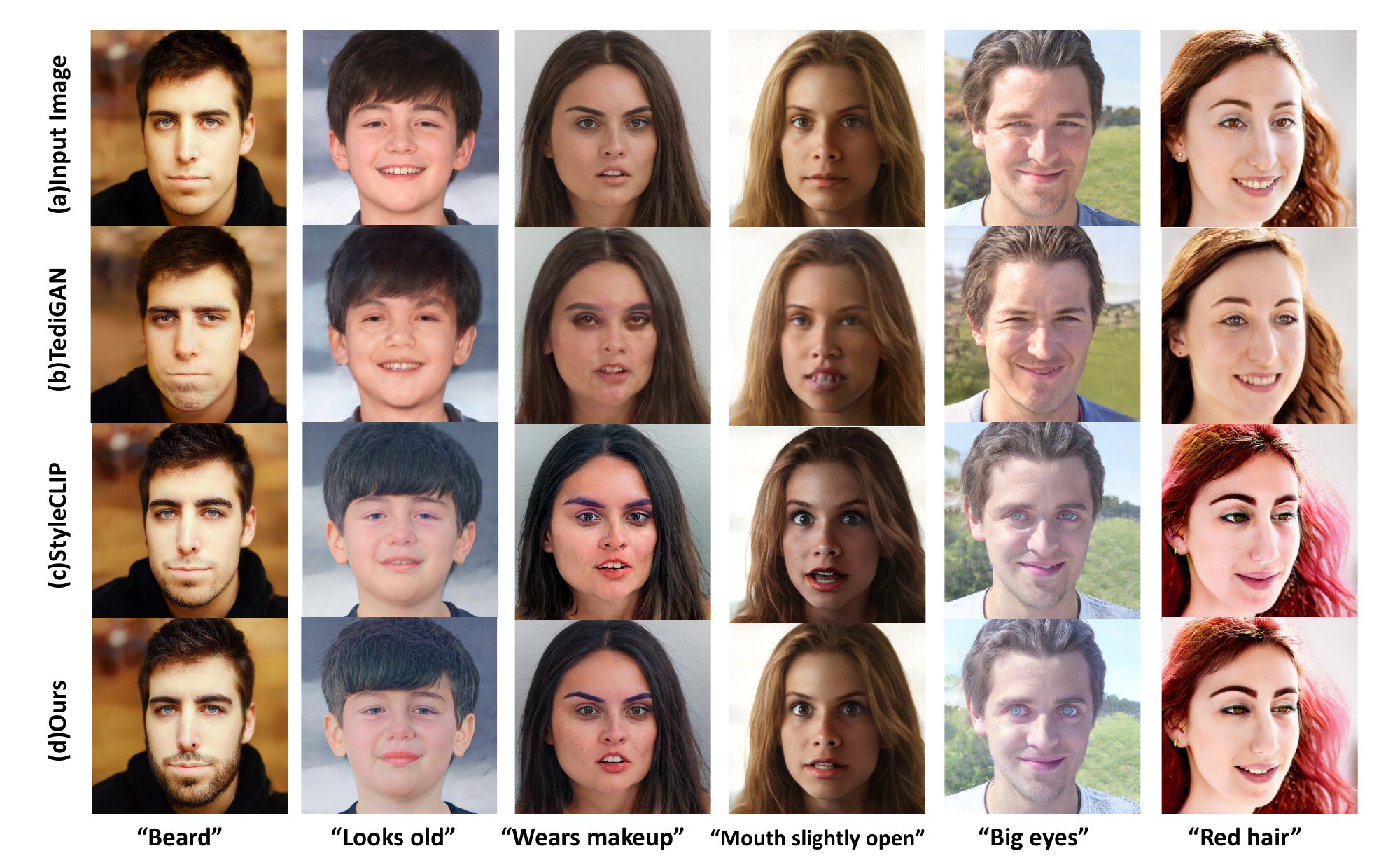}
  \caption{Qualitative comparison of text-guided image manipulation compared with the state-the-of-art methods. TextCLIP accomplishes more accurate semantic editing against the original image than previous methods.}
  \label{f6}
\end{figure*}
\textbf{User study}: we also conducted a user study. 10 users from different backgrounds were selected and a user study was conducted by randomly generating 50 images under the same textual conditions. User request to rank images generated by different models under the same conditions. The user study consisted of the following aspects:
\begin{itemize}
    \item \textbf{Image realism}: to evaluate whether the generated images are realistic.
    \item \textbf{Semantic similarity}: for the image generation task, semantic similarity refers to whether the generated image is semantically consistent with the corresponding text; for the image manipulation task, semantic similarity refers to whether the model modifies the input image according to the specified text.
\end{itemize}

\begin{table}[t]
	\caption{Quantitative Comparison of Text-Guided Image Generation on the Multi-modal CelebA-HQ dataset.}
	\label{gen-1}
	\begin{tabular}{ c | c  c  c }
	\toprule
	Method & FID$\downarrow$ & R-Precision$\uparrow $& LPIPS$\downarrow$ \\ 
	\midrule
		AttnGAN \cite{xu_attngan_2018} & 125.98 &0.232 & 0.512\\ 
		ControlGAN \cite{li2019controllable} & 116.32& 0.286& 0.522\\ 
		DFGAN \cite{tao2020df} & 137.60&0.343 & 0.581\\ 
		DM-GAN \cite{zhu_dm-gan_2019} &131.05 & 0.313& 0.544\\ 
		TediGAN \cite{xia2021tedigan} & 106.37 &0.188 & 0.456\\ 
		\textbf{TextCLIP (ours)} & \textbf{88.27} & \textbf{0.384} & \textbf{0.396}\\ 
		\bottomrule 
	\end{tabular}
\end{table}

\begin{table}[t]
	\caption{User Study on Multi-modal CelebA-HQ dataset. Acc. denotes semantic similarity and Real. denotes image realism.}
	\label{gen-2}
	\begin{tabular}{ c | c  c  c }
		\toprule
		Method & Acc. (\%)$\uparrow$ & Real.(\%)$\uparrow$ \\ 
		\midrule
		AttnGAN \cite{xu_attngan_2018} & 18.6 & 12.8 \\ 
		ControlGAN \cite{li2019controllable} & 19.7& 13.9\\ 
		DM-GAN \cite{zhu_dm-gan_2019} & 21.1 &16.3\\
		TediGAN \cite{xia2021tedigan} & 17.8 &22.3\\ 
		\textbf{TextCLIP (ours)} & \textbf{27.8} & \textbf{39.7} \\ 
		\bottomrule
	\end{tabular}
\end{table}
\subsection{Results on Text-Guided Image Generation}
\subsubsection{\textbf{Quantitative Results}}
As shown in Table \ref{gen-1}, on the Multi-Modal CelebA-HQ Dataset \cite{xia2021tedigan}, we compared the three metrics FID, LPIPS, R-precision with previous works. Based on the powerful image generation capability of StyleGAN \cite{karras_analyzing_2020} and the powerful image text representation capability of CLIP \cite{radford2021learning}, our proposed TextCLIP surpasses the previous state-of-the-art approachs. Our proposed level-channel mapper can map textual information to the latent space $\mathcal{W}+$ of StyleGAN well and achieve high-quality image generation. At the same time, the loss function we designed can ensure generate the clearest possible images while ensuring semantic alignment. As shown in Table \ref{gen-2}, user research shows that our approach outperforms the previous state-of-the-art approaches in terms of image realism and semantic similarity.
\begin{table}[t]
	\caption{Quantitative Comparison and User Study of Text-Guided Image Manipulation on the Multi-modal CelebA-HQ Dataset. Acc. denotes semantic similarity and Real. denotes image realism.}
	\label{t4}
	\begin{tabular}{ c | c  c  c c }
		\toprule
		Method  & IDS$\uparrow $ &  LPIPS $\downarrow$  & Acc.(\%)$\uparrow$ &  Real.(\%)$\uparrow $ \\ 
		\midrule
		TediGAN \cite{xia2021tedigan} &0.18& 0.45 & 10.8&12.4 \\ 
		StyleCLIP \cite{patashnik2021styleclip} & 0.76&0.42& 38.9&40.1\\ 
		\textbf{TextCLIP (ours)}&\textbf{0.84} &\textbf{0.39}& \textbf{50.3} & \textbf{47.5} \\ 
		\bottomrule 
	\end{tabular}
\end{table}
\subsubsection{\textbf{Qualitative Results}}
As shown in Figure \ref{f5}, we compare qualitatively with the previous state-of-the-art methods. The comparison shows that our generated images have higher semantic similarity and image fidelity. In terms of semantic similarity, we use the semantic loss for supervision and exploit the powerful cross-modal text-image representation capability of the CLIP model to achieve higher cross-modal semantic alignment compared to other methods. In terms of image fidelity, we generated more realistic and realistic, higher resolution images. Unlike previous studies, we introduced an image fidelity loss to ensure that the generated images are realistic enough, taking into account the model's overfitting to semantic loss. Also based on the powerful generative power of StyleGAN, images with a resolution of $1024 \times 1024$ were generated. While AttnGAN \cite{xu_attngan_2018} and ControlGAN \cite{li2019controllable} only can generate lower resolution images and TediGAN \cite{xia2021tedigan} sometimes generates some blurred images. Take the sentence "She has a pointy nose with her mouth closed" as an example, the focus is on "she", "pointy nose" and "mouth closed".Our generated images are highly semantically aligned with these three features; whereas TediGAN generated images with mouths not closed, AttnGAN and ControlGAN generated somewhat blurred and low reslution images.As shown in Figure \ref{ff2},for the same text, our method generates several different images, which demonstrates the diversity of our text-guided image generation methods.

\subsection{Results on Text-Guided Image Manipulation}
\subsubsection{\textbf{Quantitative Results}}
As shown in Table \ref{t4}, we compared with the previous TediGAN \cite{xia2021tedigan}, StyleCLIP \cite{patashnik2021styleclip}. Instead of using FID to evaluate text-guided image manipulation as in previous methods, we use IDS to evaluate whether the identity information is well preserved before and after the image is semantically modified,and use LPIPS to determine whether some semantically irrelevant image regions are preserved. And we conduct user study to determine the goodness of the model. The experiments show that, in contrast to previous methods, our proposed TextCLIP does a good job of semantically editing relevant image regions and partially preserving irrelevant image regions.
\subsubsection{\textbf{Qualitative Results}}
As shown in Figure \ref{f6}, we compare it with the previous TediGAN \cite{xia2021tedigan}, StyleCLIP \cite{patashnik2021styleclip}. Our method does a good job of modifying the semantically relevant parts according to the specified text, while not modifying the semantically irrelevant parts. In all six examples, TediGAN does not generate semantically relevant images well, while StyleCLIP produces similar results to our method, but the images produced by our method are more relevant to the given text while retaining the semantically irrelevant image regions well. This is not only because our designed level-channel mapper accurately maps the initial latent code according to the conditions of corresponding text, but also because our designed loss functions, including identity loss and semantic loss, accurately modify the images, preserving semantically irrelevant regions of the images such as face identity well.

\section{Ablation Study}
\subsection{Ablation Study On Loss Functions}
 As shown in Table \ref{t5}, we designed a loss function that helps to improve the performance of the text-guided image generation and manipulation tasks. The semantic loss function makes the generated images semantically consistent with the given text, which takes advantage of CLIP \cite{radford2021learning} strong image-text representation capability. The identity loss function, especially on text-guided image manipulation tasks, allows for the good preservation of identity information of face images. The image loss function and the fidelity function allow the generated image to be close to the original image while being more realistic.
\subsection{Ablation Study On Network Structures}
As shown in Table \ref{t6}, the level-channel mapper demonstrates a powerful performance combined with StyleGAN \cite{karras_analyzing_2020} and CLIP \cite{radford2021learning}. The level mapper helps to extract features in a hierarchical manner, and the channel mapper enables finer control of text-based conditions at a finer granularity. The experimental results show that the level-channel mapper formed by the combination of level mapper and channel mapper has excellent performance.

\begin{table}[t]
	\caption{Ablation Study On Loss Function. Gen. denotes image generation, Man. denotes image manipulation.}
	\label{t5}
    \begin{tabular}{l|cc|cc}
	\toprule
	\multirow{2}{*}{Method} &
    \multicolumn{2}{c|}{Gen.}&
    \multicolumn{2}{c}{Man.} \\
     & FID$\downarrow $& R-precision$\uparrow$ & IDS$\uparrow$ & LPIPS$\downarrow$\\
    \midrule
    w/o $\mathcal{L}_{semantic}$ &  99.93 & 0.143 & 0.11& 0.46\\
    w/o $\mathcal{L}_{ID}$  & 90.34 & 0.433 & 0.34 & 0.44\\
    w/o $\mathcal{L}_{img}$  & 94.54 & 0.428 & 0.78 & 0.45\\
    w/o $\mathcal{L}_{d}$  & 93.28 & 0.483 & 0.83 & 0.40\\
    \midrule
    \textbf{TextCLIP (ours)}  & \textbf{88.27}&\textbf{0.384} & \textbf{0.84}& \textbf{0.39}\\
    \bottomrule
\end{tabular}
\end{table}

\begin{table}[t]
	\caption{Ablation Study On Network Structure. Gen. denotes image generation, Man. denotes image manipulation.}
	\label{t6}
    \begin{tabular}{l|cc|cc}
	\toprule
	\multirow{2}{*}{Method} &
    \multicolumn{2}{c|}{Gen.}&
    \multicolumn{2}{c}{Man.} \\
     & FID$\downarrow$ & R-precision$\uparrow$ & IDS$\uparrow$ & LPIPS$\downarrow$\\
    \midrule
    w/o Level Mapper & 92.46 &0.448 & 0.81 & 0.48\\
    w/o Channel Mapper  & 100.22&0.396 & 0.78& 0.42\\
    \midrule
    \textbf{TextCLIP (ours)}  & \textbf{88.27}&\textbf{0.384} & \textbf{0.84}& \textbf{0.39}\\
    \bottomrule
\end{tabular}
\end{table}

\section{Limitations}
After analysis we believe there are several limitations:
\begin{itemize}
    \item TextCLIP is only done for specific face domains now, in the future we hope to extend this method to other domains such as flowers, birds, etc. In order to verify the superiority of the performance of our method on the flower and bird domains, we need to pre-train StyleGAN on the relevant flower and bird datasets. The StyleGAN pre-trained on the flower and bird dataset can generate high resultion flower and bird pictures, which is our next step in the future.
    \item Since TextCLIP is based on StyleGAN \cite{karras_analyzing_2020} and CLIP \cite{radford2021learning}, the problems that arise in CLIP and StyleGAN itself will also arise in TextCLIP. For example, some attributes, such as hats and earrings, are not well represented in the latent space of StyleGAN so we do not get the desired results. In addition, CLIP is at risk of being attacked.
\end{itemize}

\section{Conclusion}
Based on the powerful image generation capabilities of StyleGAN and the image text alignment capabilities of Contrastive Language-Image Pre-training(CLIP), we propose a new approach that provides a unified framework for text-guided image generation and manipulation, does not require adversarial training, and can accept open-world texts . Extended experiments on the Multi-modal CelebA-HQ dataset demonstrate that our approach outperforms previous state-of-the-art methods in both text-guided image generation tasks and text-guided image manipulation tasks. In the future, we hope that TextCLIP will not be limited to the face domain, but will be extended to other domains such as flowers, birds, etc. In addition, for text-guided image manipulation tasks, we would like to explore a unified approach which does not need to go through the process of training different models for different classes of textual conditions, using only one model to complete the task.

\bibliographystyle{ACM-Reference-Format}
\bibliography{ref.bib}










\end{document}